\documentclass{article}

\usepackage{arxiv}

\usepackage[utf8]{inputenc} 
\usepackage[T1]{fontenc}    
\usepackage[colorlinks=true, allcolors=blue]{hyperref}
\usepackage{url}            
\usepackage{booktabs}       
\usepackage{amsfonts}       
\usepackage{nicefrac}       
\usepackage{microtype}      
\usepackage{lipsum}
\usepackage{graphicx}
\graphicspath{ {./images/} }
\usepackage{authblk}
\usepackage{amsmath,amssymb, amsthm, mathtools, bbm}
\usepackage{multirow}

\theoremstyle{definition}

\title{Comparing the Effects of Persistence Barcodes Aggregation and Feature Concatenation on Medical Imaging}

\author[a]{Dashti A. Ali}
\author[b]{Richard K. G. Do}
\author[c]{William R. Jarnagin}
\author[d]{Aras T. Asaad}
\author[a,e]{Amber L. Simpson}

\affil[a]{School of Computing, Queen’s University, Kingston, ON, Canada}
\affil[b]{Department of Radiology, Memorial Sloan Kettering Cancer Center, New York, USA}
\affil[c]{Hepatopancreatobiliary Service, Department of Surgery, Memorial Sloan Kettering Cancer Center, New York, USA}
\affil[d]{Mathematical Institute, Oxford University, Oxford, United Kingdom}
\affil[e]{Department of Biomedical and Molecular Sciences, Queen’s University, Kingston, ON, Canada}

\begin{document}
\maketitle

\begin{abstract}
In medical image analysis, feature engineering plays an important role in the design and performance of machine learning models. Persistent homology (PH), from the field of topological data analysis (TDA), demonstrates robustness and stability to data perturbations and addresses the limitation from traditional feature extraction approaches where a small change in input results in a large change in feature representation. Using PH, we store persistent topological and geometrical features in the form of the persistence barcode whereby large bars represent global topological features and small bars encapsulate geometrical information of the data. When multiple barcodes are computed from 2D or 3D medical images, two approaches can be used to construct the final topological feature vector in each dimension: aggregating persistence barcodes followed by featurization or concatenating topological feature vectors derived from each barcode. In this study, we conduct a comprehensive analysis across diverse medical imaging datasets to compare the effects of the two aforementioned approaches on the performance of classification models. The results of this analysis indicate that feature concatenation preserves detailed topological information from individual barcodes, yields better classification performance and is therefore a preferred approach when conducting similar experiments.
\end{abstract}


\section{Introduction}
The development of machine learning (ML) models in medical imaging has gained significant momentum due to their efficacy in a wide variety of clinical tasks, including prediction and prognostication. Feature engineering is one of the most important components of any ML pipeline. Classical approaches, such as radiomics, rely on pixel-wise comparisons but are sensitive to variations in image acquisition settings (e.g. differences in contrast or resolution) \cite{mayerhoefer2020introduction, lee2021radiomics, rizzo2018radiomics}. Deep learning methods help address some of these challenges; however, opaque decision-making processes reduce explainability, and high computational demands hinder accessibility. Topological data analysis (TDA) has emerged as a promising mathematical technique to overcome some of these limitations, extracting additional insights that cannot be obtained by classical approaches alone.
TDA is primarily based on the field of mathematics called algebraic topology, where topology captures the shape and structure of data at different scales, revealing complex relationships and identifying hidden patterns \cite{carlsson2009topology, lum2013extracting}. Persistent Homology (PH), one of the main tools in the field of TDA, summarizes the topological and geometrical features of the data, such as connected components, loops, and cavities, in the form of persistence barcodes (PB) \cite{carlsson2009topology, ghrist2008barcodes, edelsbrunner2022computational}. Several feature vectors can be built from a PB using vectorization methods, and these features can subsequently be utilized for various ML tasks.

TDA has been successfully applied to various medical imaging modalities such as X-ray, CT, MRI, and ultrasound \cite{singh2023topological}. Some examples of these applications include explaining the geometric structure of the airway system from lung CT scans and classifying chronic obstructive pulmonary disease \cite{belchi2018lung}, analyzing osteoarthritis from MRI images \cite{pedoia2018mri}, distinguishing benign from malignant tumours and adenocarcinoma from squamous cell carcinoma in lung CT images \cite{vandaele2023topological}, predicting the progression of hepatic decompensation \cite{singh2022algebraic} or evaluating the accuracy of hepatic cancer classification \cite{oyama2019hepatic} from MRI images, and assessing tumour heterogeneity in lung adenocarcinomas from CT images \cite{kawata2021representation}, among others. A survey paper, including a comprehensive literature review on the recent applications of TDA in medical imaging, is available in \cite{singh2023topological}.

There are different methods to obtain the final set of TDA features in medical imaging. Most often, PH features are extracted from the entire 3D image, resulting in a single PB that encapsulates topological information in feature vectors. However, features acquired in this way may not always provide a robust representation of the image, which could affect the performance of the ML model. For example, with a small dataset of images, one might construct PH on each 2D slice of the image separately, which would result in the computation of multiple barcodes from a single image. In ultrasound imaging, this may occur when PH is built from each filtered version of the image. An open problem remains: should barcodes be combined by aggregation or concatenation? The aim of this study is therefore to systematically compare the model performance with these approaches evaluated with multiple datasets.

The contribution of this paper is as follows: (1) generate multiple barcodes from 3D and 2D medical images, (2) investigate the effect of barcode aggregation versus barcode-feature concatenation on final ML performance using different medical imaging modalities such as 3D images of CT scans and 2D images of ultrasounds and mammograms, and (3) implement five different PH vectorizations together with 19 different classification methods to showcase the effect of barcode aggregation versus barcode-feature concatenation on ML performance.

\section{Methods}
Extracting TDA features from medical images usually involves two steps. First, a combinatorial approach, also known as filtration (e.g. cubical complex), is employed to construct persistent homology from the image in the form of PB. Second, using the vectorization method to featurize the space of PBs, thereby obtaining feature vectors. There are different ways to build PH from images. In this study, we will use two distinct approaches: (1) using image patch local binary patterns, a landmark selection algorithm with simplicial complex filtration, and (2) cubical complex filtration.

\subsection{Image Patch Local Binary Patterns (IP-LBPs)}
\begin{figure}[h]
\begin{center}
\includegraphics[width=0.75\linewidth]{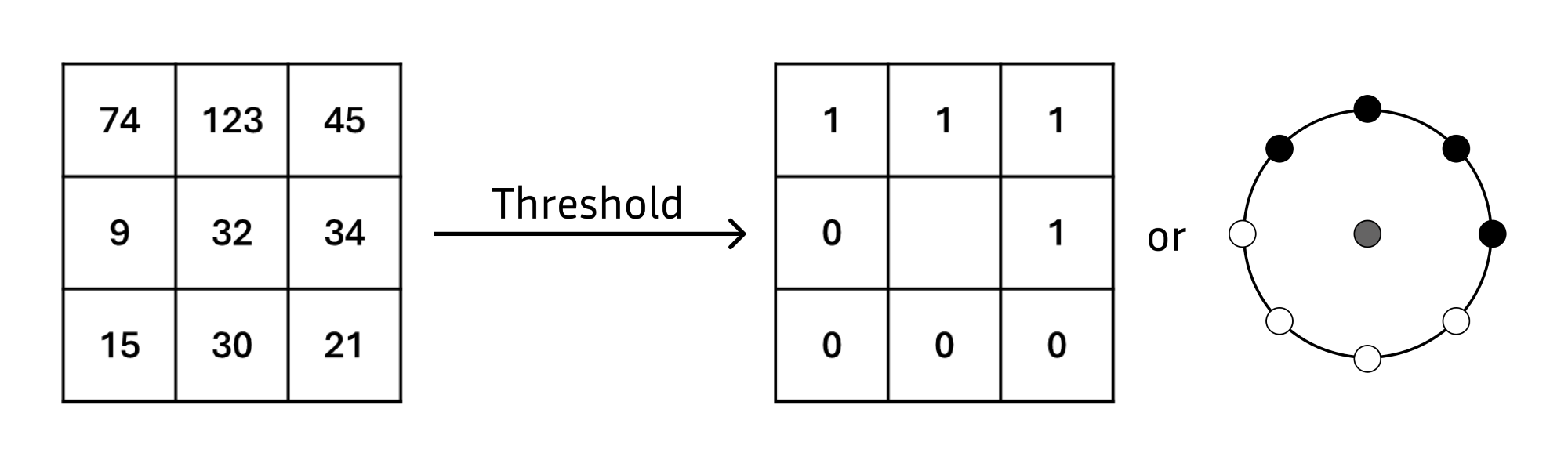}
\end{center}
\caption
{ \label{fig:ulbp_process} 
The LBP process takes patches of 3x3 pixels from the image, compares the central pixel with its neighbours, and then adds 1 to its binary cyclic 8-cell representation if the neighbouring pixel is greater than or equal to the central one, otherwise 0; this will result in a binary pattern indicating the transitions between 0 and 1.
}
\end{figure}

\begin{figure}[h]
\begin{center}
\includegraphics[width=0.9\linewidth]{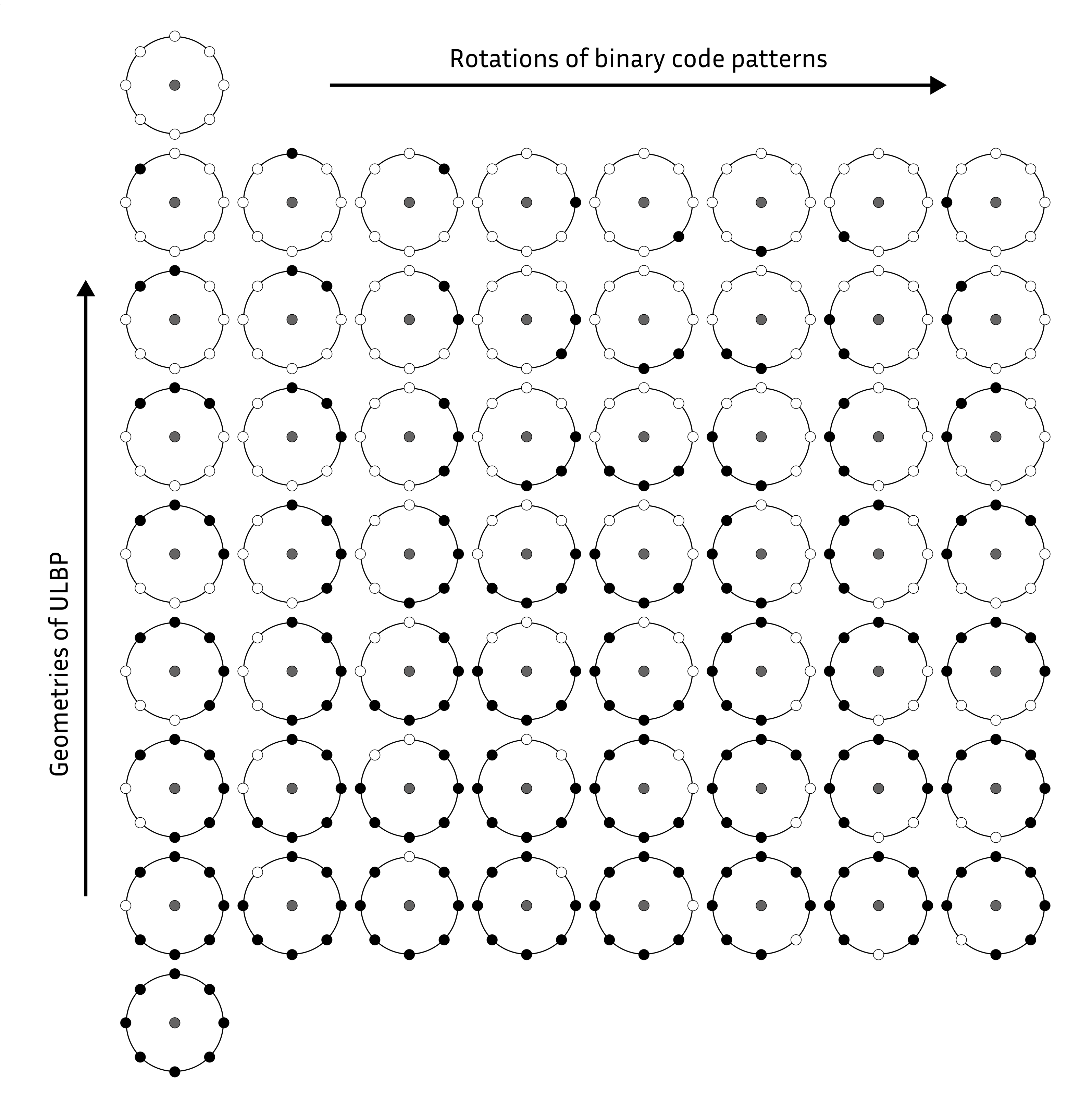}
\end{center}
\caption
{ \label{fig:ulbp_geometries}
Geometric visualization of the ULBP method.
}
\end{figure}

Local binary patterns (LBP) is a texture descriptor approach to encode the local structure and texture of images \cite{ojala1996comparative}. As illustrated in figure \ref{fig:ulbp_process}, the LBP process on an image starts by scanning it from the top-left corner, taking 3×3 pixel patches. For each patch, the central pixel's intensity is compared with its neighbours: assign 1 if the neighbour is greater than or equal to the centre, otherwise 0, forming a binary cyclic 8-cell representation. This will result in a binary pattern with transitions between 0 and 1. In order to account for the pixels at the edges, the image will be padded by zeros. There are in total 256 different possible patterns, with a subset of them called uniform LBPs (ULBPs), a set of 58 distinct binary patterns with only 0 or 2 circular transitions between 0's and 1's. This set of ULBPs can be further categorized as follows: beside the two binary patterns with 8 zeros and 8 ones, the remaining 56 binary patterns (with 2 circular transitions) can be organized into 7 groups called geometries, where each contains 8 unique variations called rotations. This categorization is illustrated geometrically in figure \ref{fig:ulbp_geometries}. As highlighted by Ojala et al. \cite{ojala2002multiresolution}, the ULBPs can represent 90\% of the patterns and textures in natural images. In this study we have used a small subset of ULBPs, including the fourth and fifth geometries only. By employing this method, one can select landmark pixels that follow a specific pattern from an input image from which a point cloud can be constructed. Selected pixel locations using this approach will be the basis of building PH barcodes in 3D and 2D images. Next, we describe simplicial and cubical complexes which are the two approaches we use to build the topology of medical images.

\subsection{Simplicial Complex}
A simplicial complex (SC) is a topological space constructed from a collection of multi-dimensional triangular subsets of $\mathbb{R}^n$, joined together along shared boundary elements like nodes, edges, and faces. Mathematically, an SC on a set of vertices $V=\{v_{0},v_{1},v_{2},\cdots,v_{n}\}$ is a collection $S$ of subsets of $V$ that satisfies the following conditions:
\begin{itemize}
\item $\{v\} \subset S$ $\forall v \subset V$
\item If \(\tau \in S\), then every subset $\sigma \subset \tau$ is also an element of $S$
\end{itemize}
Then each $\tau \in S$ is called a simplex, and its dimension is defined as $\dim(\tau) = |\tau| - 1$, where $|\tau|$ is the cardinality of $\tau$~\cite{nanda2013simplicial}.

The Vietoris–Rips $(VR)$ complex is a specific type of simplicial complex deployed in this work, defined as follows: given a set of points $\psi$ in $\mathbb{R}^2$ and a parameter $\epsilon$, the Vietoris–Rips complex $VR(\psi, \epsilon)$ is defined as the simplicial complex whose vertex set $\{v_0, v_1, v_2, \dots, v_n\}$ spans an $n$-simplex if the Euclidean distance between any two points is less than or equal to the chosen value of $\epsilon$ (i.e., $d(v_i, v_j) \leq \epsilon$ for all $0 \leq i, j \leq n$). As the value of $\epsilon$ increases, the $(VR)$ complex grows accordingly. As a result of this procedure, a nested sequence of $(VR)$ simplicial complexes, known as a filtration, is created (i.e., $VR(\psi, \epsilon_1) \subseteq VR(\psi, \epsilon_2)$ whenever $\epsilon_1 \leq \epsilon_2$).
During this process, homological features appear and vanish, where each can be stored as a bar which encapsulates the birth and death of connected components and 1D loops. A collection of such bars is stored in what is called persistence barcode (PB). An equivalent representation of PB is known as persistence diagram (PD) in which a bar is represented as a point in 2D above the off-diagonal line $y=x$ \cite{carlsson2009topology, nanda2013simplicial, otter2017roadmap}.

Next, we describe cubical complex as a second method to build persistence barcodes from digital images, which does not require an image landmark selection step. Rather, it uses the grid structure of the medical images to keep track of connected pixels/voxels and loops that appear and disappear when using gray-level pixel intensity as a filtration parameter.
\subsection{Cubical Complex}
Cubical complex, the cubical analogue of a $(VR)$ simplicial complex, is a method to build PH from grid-structured data, for instance, digital images. Using a scalar function, gray-level pixel intensity here, cubical complex is constructed by ordering the cells (vertices, edges, squares, etc.). For example, for a digital image, cells are added to the complex based on the ascending order of their pixel intensity values.
A finite cubical complex in \(\mathbb{R}^{2}\) is a collection of cubes aligned on the grid \(G^{2}\), meeting some conditions analogous to a simplicial complex \cite{garin2019topological}. A \(2\)-dimensional image can be written as a map \(\eta:I \subseteq G^{2} \rightarrow \mathbb{R}\). For an element \(v\in I\), which is referred to as a voxel (pixel when \(d=2\)), its intensity or grayscale value is denoted as \(\eta(v)\). Digital images can be represented as cubical complexes in various ways. A grayscale image comes with a natural filtration within the grayscale values of its pixels. Voxels can be represented by vertices, and cubes are formed between them. Extending the values of voxels to all the cubes \(\omega \in C\), a function on the resulting cubical complex \(C\) is obtained as follows:
\[
\eta'(\omega) \coloneqq \min_{\tau \text{ (face of } \omega)} \eta(\tau).
\]
Assume \(C\) to be the cubical complex constructed on the image \(I\) and let 
\[
C_{i} \coloneqq \{\omega \in C \mid \eta'(\omega)\leq i\},
\]
be the \(i\)-th sublevel set of \(C\). Then a filtration of the cubical complex, indexed by the values of the function \(\eta\), can be defined as \(\{C_{i} \}_{i\in Im(I)}\) \cite{garin2019topological}.

\begin{figure}[h]
\begin{center}
\includegraphics[width=1\linewidth]{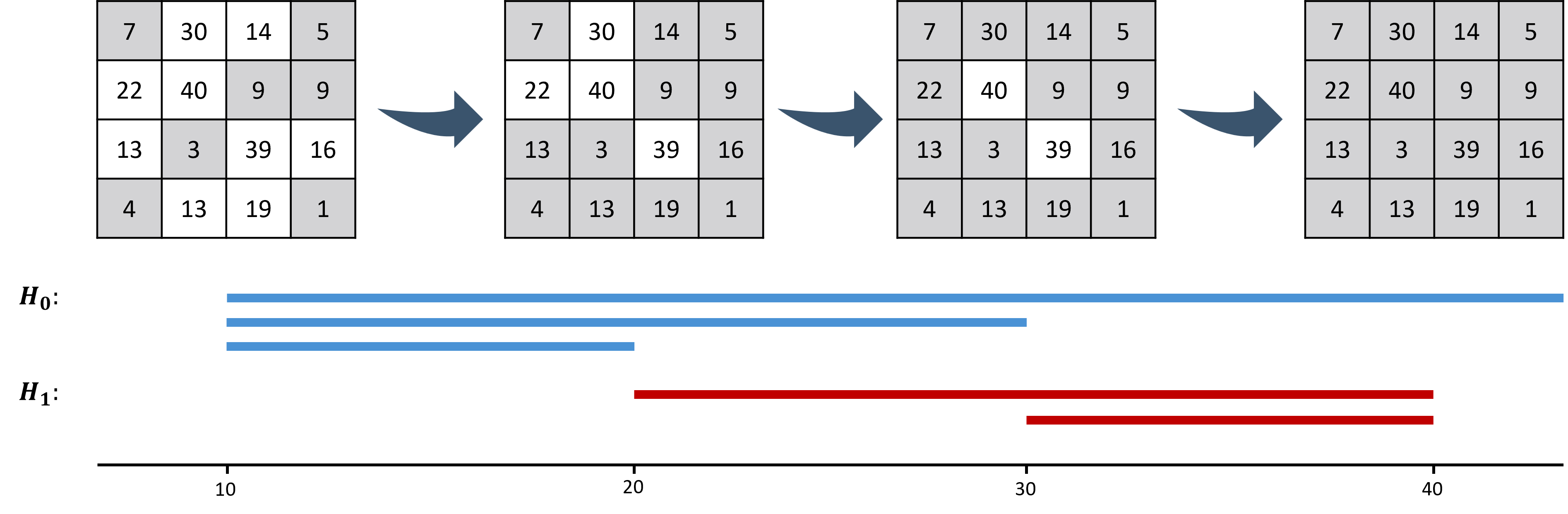}
\end{center}
\caption
{ \label{fig:cubical_complex_filtration} 
Cubical complex filtration and persistence barcodes representations in dimension zero and one for a 4x4 grayscale image patch.
}
\end{figure}

Figure \ref{fig:cubical_complex_filtration} explains this process on a 4x4 patch of a grayscale image. As the filtration value is increased, pixels with higher intensity are filtered and added to the complex. Simultaneously, the construction of PBs is shown in the same figure, below the grid process. 
\(H_0\) and \(H_1\) highlight the appearance (birth) and vanishing (death) of connected components and loops, respectively.

A barcode can be expressed as
\[
B = [(p_1,q_1),(p_2,q_2),\cdots,(p_n,q_n)],
\]
where \(B(p,q)\) indicates a bar with birth \(p\) and death \(q\). It is important to note that a barcode has a multiset structure, meaning an element $(p, q)$ in the barcode $B$ can appear multiple times \cite{hofer2019learning}. As a result, the barcode $B$ is associated with a multiplicity function $\mu: B \rightarrow \mathbb{Z}_{>0}$.

\subsection{Persistence Barcode Vectorization and Aggregation}
\begin{figure}[h]
\begin{center}
\includegraphics[width=1\linewidth]{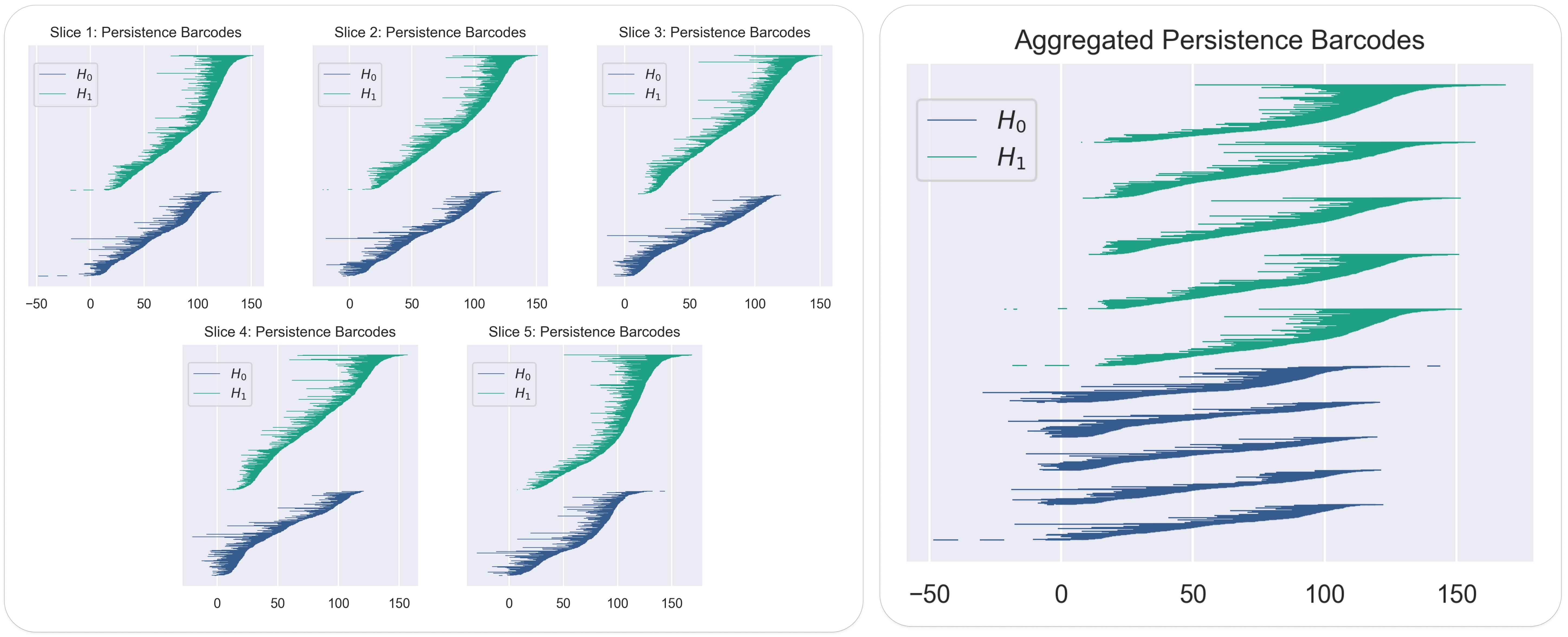}
\end{center}
\caption
{ \label{fig:single_vs_aggregated_barcodes} 
\textbf{Left}: persistence barcodes per slice, \textbf{Right}: aggregated persistence barcodes of all slices.
}
\end{figure}

Topological and geometrical features stored in persistence barcodes are intervals encapsulating the birth and death of such features across many distance/similarity resolutions. when more than one persistence barcode generated from a single 2D image or a sequence of PBs are generated from slices of CT scan of a patient, then one can either (1) aggregate PBs into a single PB and then vectorize it or (2) vectorize each PB using a suitable vectorization method and then concatenate vectorized barcodes into a single feature vector. Barcode aggregation means simply combine all intervals of one PB with another PB of the same dimension keeping all duplicates and the original order. In figure \ref{fig:single_vs_aggregated_barcodes} we show a group of persistence barcodes in dimension 0 and 1 and their corresponding aggregated version plotted in the right. 
The topological features stored in persistence barcodes are not compatible with ML algorithms because their Fréchet averages are not unique. Hence, barcodes need to be embedded into vector spaces such as Hilbert (or Banach) spaces, where well-defined inner products are possible.
There have been several vectorization techniques developed in the last 10 years to embed barcodes into feature vectors to be utilized in a ML pipeline \cite{chung2022persistence, chevyrev2018persistence, berry2020functional, chintakunta2015entropy}. In this work, we experiment with five different vectorization algorithms, including the Betti curve, persistent statistics, entropy summary, persistent landscapes, and persistent tropical coordinates (P.T.C.), together with the two filtration approaches of landmark-based $VR$ complex and cubical complexes. Next, we briefly describe each of the five vectorization methods we use in this work.

\textbf{Betti curve:} This widely used and simple technique transforms a PB into a feature vector by counting the number of bars in a PB that intersect each of a set of evenly spaced vertical lines $v=1,2,...,\gamma$. In this study, we choose the parameter $\gamma=100$, and consequently, a feature vector of length $\gamma$ will be acquired.

\textbf{Persistent Statistics:} In this approach, a feature vector is obtained by computing various statistical measures from a given barcode. These include the standard deviation, mean, median, full range, interquartile range, and the 10th, 25th, 75th, and 90th percentiles of the birth $p$, the death $q$, the midpoints, and the lifespans $q-p$ of all the bar intervals $[p, q]$ in a PB. This set of features is further extended to include bar counts and entropy \cite{chintakunta2015entropy}.

\textbf{Entropy Summary:} The entropy summary function is a piecewise constant function introduced by Atienza et al. \cite{atienza2020stability}. For a barcode $B$ with its multiplicity function $\mu$, the entropy summary function of $\mu$ is the mapping $S_\mu:\mathbb{R}\rightarrow \mathbb{R}$ given by:
\[
S_\mu(t)=-\sum_{[p,q]\in B} \mathbbm{1}_{p\leq t<q} \cdot \mu_{p,q} \cdot \left( \frac{q-p}{L_\mu} \right) \cdot \log\left( \frac{q-p}{L_\mu} \right),
\]
where $\mathbbm{1}_\bullet$ is the indicator function (equal to 1 if $\bullet$ is true and 0 otherwise) and $L_\mu$ is defined as:
\[
L_\mu \coloneqq \sum_{[p,q]\in B} \mu_{p,q} \cdot \left(q-p\right).
\]

\textbf{Persistent Landscapes:} Persistent landscapes (PL), a commonly adopted vectorization technique, were introduced to map PDs into a stable and invertible function space \cite{bubenik2015statistical}. The PL of a barcode is defined as a sequence of functions $\{\Psi^\mu_i: \mathbb{R}\rightarrow \mathbb{R} \mid i\in \mathbb{Z}_{>0}\}$, where for any real number $t$ in $\mathbb{R}$, $\Psi^\mu_i(t)$ is given by:
\[
\Psi^\mu_i(t) = \sup \left\{ s \geq 0 ~ \Big{|} ~ \left(\sum_{[p,q] \in B} \mathbbm{1}_{[t-s,t+s] \subset [p,q]} \cdot \mu_{p,q}\right) \geq i\right\}.
\]

\textbf{Persistent Tropical Coordinates:} To define this method, an arbitrary ordering of the intervals in $B$ is used, denoted as $\{[p_i,q_i] \mid 1\leq i \leq n\}$, where each interval $[p,q]$ appears $\mu_{p,q}$ times. The Persistent Tropical Coordinates (PTC) of a PB are defined as a tropical and symmetric function $F(x_1,y_1,\cdots, x_n,y_n)$. The term tropical refers to the fact that $F$ is expressed using only the operations of maximum, minimum, addition, and subtraction on the variables $\{x_i\}$ and $\{y_i\}$. Symmetry means that $F$ remains unchanged under any permutation of the indices $\{1,\cdots,n\}$ applied to $\{x_i\}$ and $\{y_i\}$. PTC features are computed by evaluating the tropical coordinate functions at $x_i=\lambda_i$ and $y_i$ is set to either max$(r\lambda_i,p_i)$ or min$(r\lambda_i,p_i)$, where $q_i-p_i$ is the lifespan of the $i$-th interval in $B$ and $r\in \mathbb{Z^+}$. 
Therefore, the set of seven tropical coordinate features \cite{kalivsnik2019tropical} are defined as follows:
\begin{align*}
    F_1 &= \max_i\,\lambda_i \\ 
    F_2 &= \max_{i<j}(\lambda_i+\lambda_j) \\
    F_3 &=\max_{i<j<k}(\lambda_i+\lambda_j + \lambda_k) \\
    F_4 &= \max_{i<j<k<l}(\lambda_i+\lambda_j+\lambda_k+\lambda_l) \\
    F_5 &=\sum_i \lambda_i \\
    F_6 &=\sum_i \min(r\lambda_i,p_i) \\
    F_7 &=\sum_j \left[\max_i\big(\min(r \lambda_i,p_i)+\lambda_i\big)-(\min(r \lambda_j,p_j)+\lambda_j)\right].
\end{align*}

\begin{figure}[h]
\begin{center}
\includegraphics[width=1\linewidth]{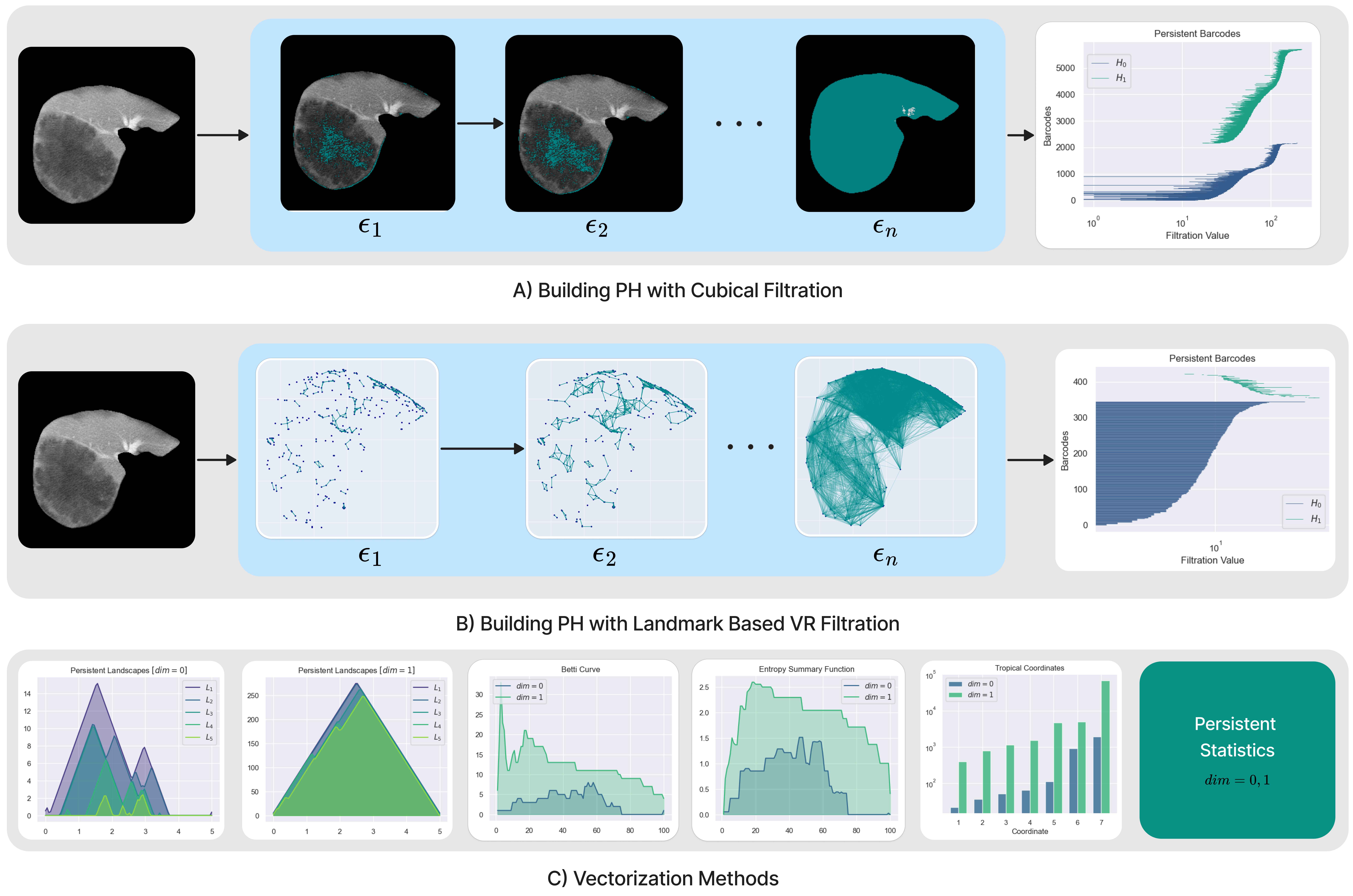}
\end{center}
\caption
{ \label{fig:TDA_pipeline} 
TDA pipeline, which includes A) PH construction using cubical complex filtration, B) PH construction using landmark-based $VR$ complex filtration and C) Various vectorization methods utilized in this study.
}
\end{figure}

We direct interested readers to see \cite{ali2023survey}, a survey on vectorization methods that covers detailed mathematical explanations of each of the five vectorization methods used in this paper and comparisons with more than half a dozen other vectorization methods in literature. Figure \ref{fig:TDA_pipeline} illustrates our TDA pipeline, including PD construction using both cubical and landmark-based $VR$ filtration and the five vectorization methods utilized in this study.

\subsection{Datasets}
\begin{figure} [ht]
\begin{center}
\includegraphics[width=1\linewidth]{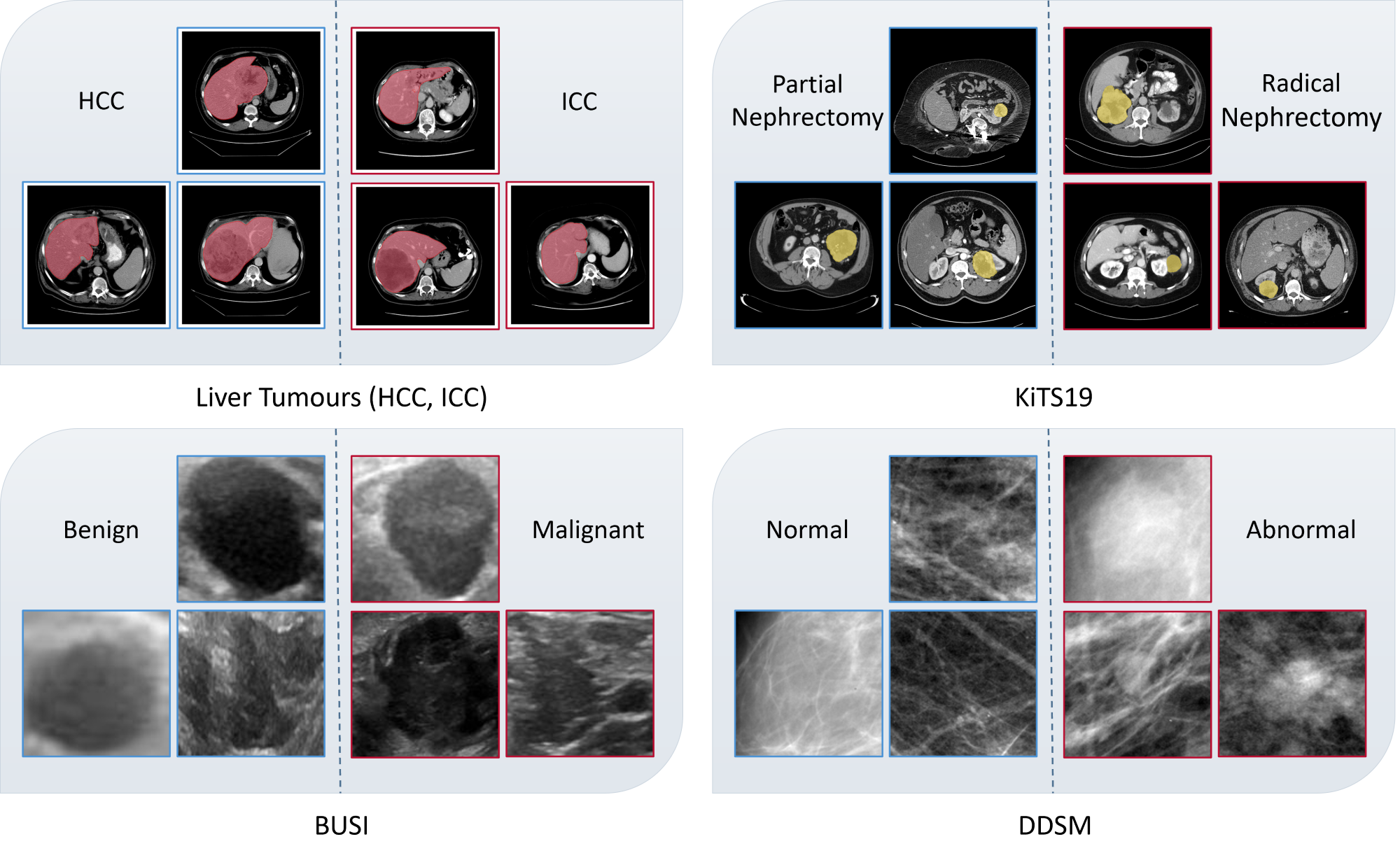}
\end{center}
\caption
{ \label{fig:dataset_sample} 
Top: sample slices of CT images from 3D datasets of liver tumours and KiTS19 with ROI highlighted in red and yellow, respectively. Bottom: sample images of ROI from the 2D datasets BUSI and DDSM. Sample images of unique classes of each dataset are separated by the dashed vertical line.
}
\end{figure}

In this work we have experimented with various datasets of 2D and 3D medical images. For the 3D images we mainly focused on computed tomography (CT) imaging modality, where an in-house dataset of liver tumours CT scans and the publicly available dataset of kidney tumours KiTS19 \cite{heller2019kits19} were considered for the experiments. The liver tumours dataset includes a balanced subset of 204 3D CT images of patients undergoing resection at Memorial Sloan Kettering Cancer Centre. These patients were diagnosed with either intrahepatic cholangiocarcinoma (ICC) or hepatocellular carcinoma (HCC). 102 CT images of each class were included in the analysis. The aim of the analysis on this dataset is to distinguish HCC from ICC. The second dataset, KiTS19, is acquired from the GitHub repository of the kidney tumours segmentation challenge, which includes CT scans of 210 patients with kidney tumours along with their segmentation masks. Seventy patients went through radical nephrectomy, while 140 proceeded with partial nephrectomy. These surgical procedures will be used as labels in downstream analysis. Sample CT slices of unique classes of each dataset are depicted in figure \ref{fig:dataset_sample} with ROI highlighted.

In the case of 2D images, two publicly available datasets were used: the breast ultrasound images (BUSI) dataset \cite{al2020dataset} and the digital database for screening mammography (DDSM) \cite{heath1998current}. A balanced subset of the BUSI dataset, which includes ultrasound images of 376 patients with benign or malignant breast tumours is included in this study whereby the analysis on this dataset aims at classifying the two types of tumours. For the DDSM dataset, 512 mammography images of patients with normal and abnormal breast tissues are considered for downstream analysis, which targets distinguishing normal from abnormal tissues. Figure \ref{fig:dataset_sample} illustrates sample images of ROI of different classes for each dataset. Table \ref{tab:datsets_summary} presents a summary of key details about each dataset. 

\begin{table}[]
\caption{Summary of Key Details for Each Dataset Included in This Study.} 
\label{tab:datsets_summary}
\begin{center}  
\begin{tabular}{|l|l|l|l|l|l|}
\hline
Dataset                                                            & Type & ROI                & Modality   & Size & Unique Classes                \\ \hline
\begin{tabular}[c]{@{}l@{}}Liver Tumours\end{tabular} & 3D   & Liver \&   Tumours & CT Scan    & 204  & HCC, ICC                      \\ \hline
KiTS19                                                             & 3D   & Tumours            & CT Scan    & 210  & Partial / Radical Nephrectomy \\ \hline
BUSI                                                               & 2D   & Tumours             & Ultrasound & 362  & Benign, Malignant             \\ \hline
DDSM                                                               & 2D   & Tumours            & Mammogram  & 512  & Normal, Abnormal              \\ \hline
\end{tabular}
\end{center}
\end{table}

\subsection{Experiment Design and Implementation}
\subsubsection{Data Preprocessing}
It is customary to have CT images of variable voxel sizes across a dataset, and our datasets of CT images are no exception. For this reason, we have resampled all CT images with average spacing across each dataset. For the liver tumours dataset, segmentation masks were utilized to crop the desired ROI, which consists of the entire liver region, including the tumours, from each CT image. The top 15 slices with the largest ROI were selected from each 3D scan where the ROI in each slice could fit in a window of 300x300 pixels, which includes sufficient padding from all sides. A similar process applied to the dataset of kidney tumours, with the ROI being the masked-out tumour regions only. A minimum of 5 non-empty slices of ROI were selected as inclusion criteria, where 7 patients were excluded as their CT images did not meet this condition. A window of 220x220 pixels used in this work which accommodates ROIs of each one of the top 5 slices of all CT images. Therefore, our final preprocessed dataset includes 5 slices of 220x220 pixels for CT scans of 203 patients. No windowing was applied on both datasets, and original Hounsfield Unit (HU) values were used.

The 2D datasets, BUSI and DDSM, consist of 128x128 pixel images of ROI. No further preprocessing was performed on these datasets.

\subsubsection{Experimental setup}
\begin{figure} [ht]
\begin{center}
\includegraphics[width=1\linewidth]{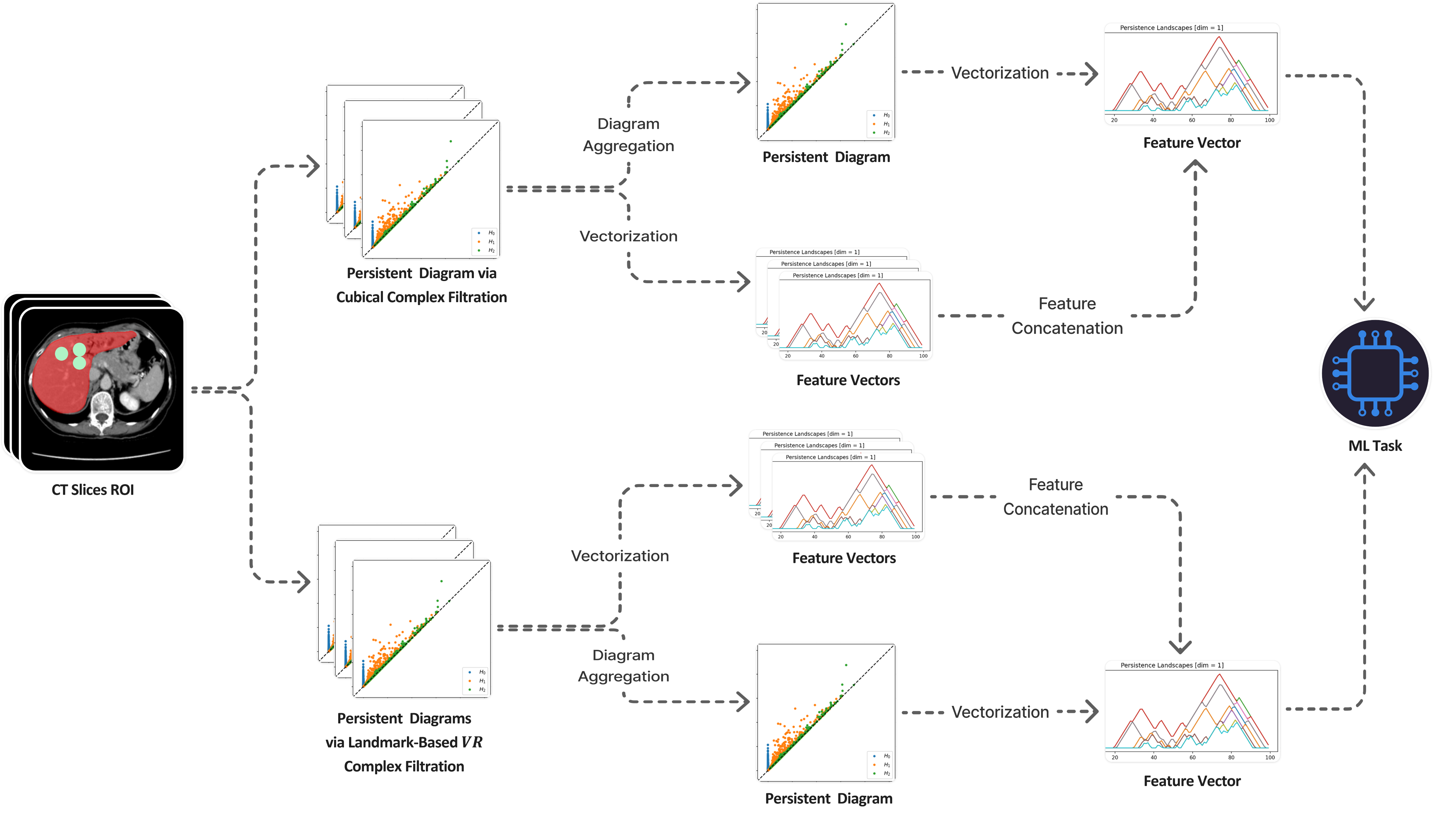}
\end{center}
\caption
{ \label{fig:3D_datasets_pipeline} 
An overview of different experiment pipelines for datasets of 3D images.
}
\end{figure}

Feature extraction is one of the main components in a classification pipeline. When it comes to PH-based features, the first step is to construct PH from the input data. In this study, we only computed PH up to dimension one. In other words, connected components and loops. For simplicity, from now on, a PD is a diagram that contains topological features of dimension zero and one. For the 3D datasets, two distinct approaches were utilized to construct PH from selected 2D slices of each CT image. In the first stream, the cubical complex filtration method was used to compute PH from each 2D slice. In the second stream, the first rotation of the fourth geometry ($G_4R_1$) of ULBP was used to select landmark points and create a point cloud from each CT slice, followed by PH construction using $VR$ complex filtration. In both cases we obtained 15 different PDs corresponding to the 15 selected slices of each CT image from the liver tumours dataset. For every CT scan in the KiTS19 dataset, only 5 PDs were acquired with each stream.
In both scenarios, barcodes can either be aggregated and then vectorized or vectorized individually first and then concatenated into a single feature vector. Therefore, a single feature vector is obtained for each PH dimension, and concatenating the two, i.e. vectorized PBs that correspond to connected components and loops, will provide the final set of features for further analysis and to be used with the ML pipeline. This overall pipeline is depicted in figure \ref{fig:3D_datasets_pipeline}.

\begin{figure} [ht]
\begin{center}
\includegraphics[width=1\linewidth]{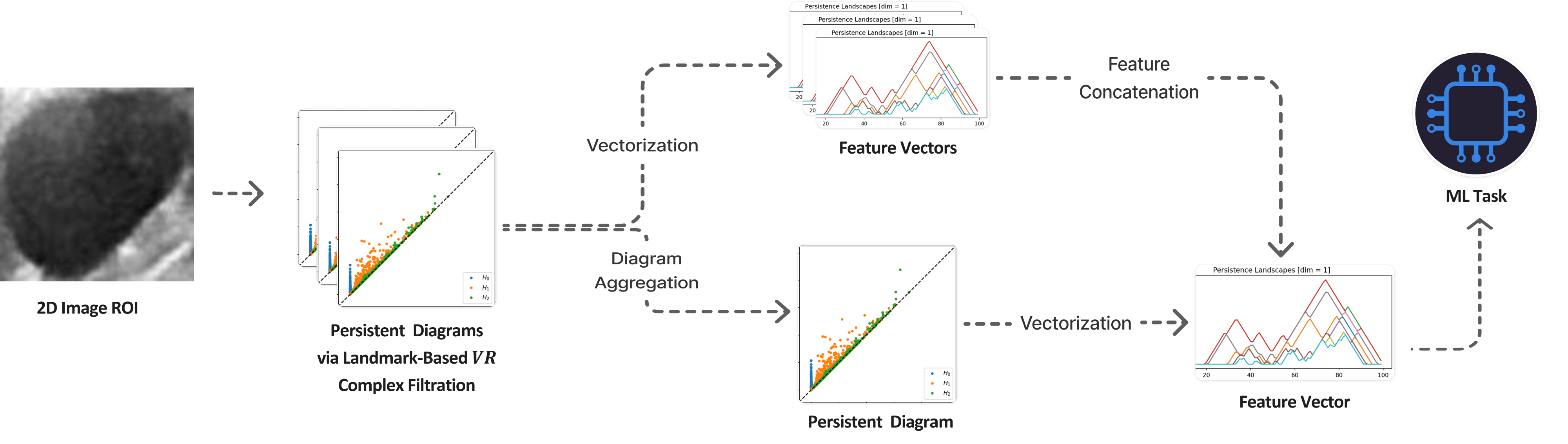}
\end{center}
\caption
{ \label{fig:2D_datasets_pipeline} 
An overview of different experiment pipelines for datasets of 2D images.
}
\end{figure}

In the case of the 2D datasets, all eight rotations of the fourth and fifth geometries of ULBP were used to select landmark points and create point clouds from each 2D image, followed by PH construction using $VR$ complex filtration. Therefore, we obtained 16 PDs for a single image. Similar to the 3D CT scan pipeline, with multiple barcodes, we performed barcode aggregation and feature concatenation to obtain a single topological feature vector for each PH dimension. Then, by concatenating these two feature vectors, a single feature vector is obtained for each 2D image with each one of the five vectorization methods. The remaining steps of the pipeline follow the same procedure as in the 3D datasets. Figure \ref{fig:2D_datasets_pipeline} illustrates this process for 2D images of BUSI and DDSM datasets.

With our final set of features from each pipeline, the next steps of the experimental setup include data splitting, standardization, feature selection and classification.
Firstly, we select 20\% of the data from each dataset randomly as an external unseen test data to measure the performance of the final ML model optimized based on the 80\% training data. This process is consistently applied on each of the individual datasets used in this work.
Furthermore, training feature vectors were standardized using z-score and feature selection was performed on the training data with absolute shrinkage and selection operator (LASSO) approach utilizing the scikit-learn package \cite{scikit-learn}. This is followed by training different classifiers on the selected features from the training data and drawing the final prediction on the test data. The classification task was performed with the PyCaret package \cite{PyCaret} using over a dozen classifiers, which are listed in Table \ref{tab:list_of_classifiers}. First, a 5-folds cross validation experiment was performed on the training set with 19 different classifiers. Then, results were sorted with respect to the average 5-folds accuracy and only the top three best performing classifiers were selected to be fine-tuned with the Optuna package \cite{akiba2019optuna}. This is followed by constructing and fine-tuning a stacking ensemble model from the selected best three models. Finally, the fine-tuned three best models alongside the ensemble one were evaluated on the testing dataset. The Python implementation source code of the experiments conducted in this project is available through this GitHub repository
\footnote{\url{https://github.com/dashtiali/barcode_aggregation_vs_feature_concatenation}}.

\begin{table}
\caption{The List of Classifiers Included in PyCaret Package and Considered in This Study.}
\label{tab:list_of_classifiers}
\begin{center}
\begin{tabular}{|c|c|}
\hline
Short Name & Name                            \\ \hline
LR         & Logistic Regression             \\ \hline
KNN        & K-Nearest Neighbors Classifier          \\ \hline
NB         & Naive Bayes                     \\ \hline
DT         & Decision Tree Classifier        \\ \hline
SVM        & SVM - Linear Kernel             \\ \hline
RBFSVM     & SVM - Radial Kernel             \\ \hline
GPC        & Gaussian Process Classifier     \\ \hline
MLP        & Multi-layer Perceptron Classifier                  \\ \hline
Ridge      & Ridge Classifier                \\ \hline
RF         & Random Forest Classifier        \\ \hline
QDA        & Quadratic Discriminant Analysis \\ \hline
ADA        & Ada Boost Classifier            \\ \hline
GBC        & Gradient Boosting Classifier    \\ \hline
LDA        & Linear Discriminant Analysis    \\ \hline
ET         & Extra Trees Classifier          \\ \hline
XGBoost    & Extreme Gradient Boosting       \\ \hline
LightGBM   & Light Gradient Boosting Machine \\ \hline
CatBoost   & CatBoost Classifier             \\ \hline
Dummy      & Dummy Classifier                \\ \hline
\end{tabular}
\end{center}
\end{table}

\section{Results}
In this study we report relevant evaluation metrics such as AUC, accuracy, recall, precision, and F1 score to benchmark the performance of each approach. Table \ref{tab:test_matrics_3D_datasets} reports the performance of each approach on the testing data of the 3D CT scan datasets, including liver tumours and KiTS19, with different filtration methods. The classification results of the best performing classifier, among 19 classifiers, and best vectorization method, among five vectorization algorithms, are reported with each one of the two streams: barcode aggregation and feature concatenation. Detailed performance of the vectorization methods and classifiers can be seen in the supplementary tables. The feature concatenation method produces a significantly higher performance with betti curve features trained with light gradient boosting classifier (LightGBM) in terms of all the metrics when compared to the barcode aggregation strategy. Similar trends can be observed in the table where barcodes are generated using landmark-based $VR$ complex filtration. Again, feature concatenation outperforms barcode aggregation in terms of all the metrics except AUC. Overall, our experiments on the liver tumours dataset suggest that the first approach yields better results compared to the latter one. Further experimental results on the liver tumour dataset, including the results of the best performing models with all five vectorization methods, can be found in tables \ref{tab:test_matrics_HCC_ICC_dataset_cubical_conplex} and \ref{tab:test_matrics_HCC_ICC_dataset_ULBP}.

\begin{table}[!h]
\caption{Comparison of Barcode Aggregation and Feature Concatenation Approach Based on Their Best Performing Classifier and Feature Type on the Testing Data of 3D Datasets Using Cubical Complex Filtration and Landmark-Based $VR$ Complex Filtration.}
\label{tab:test_matrics_3D_datasets}
\begin{center}
\begin{tabular}{|c|c|c|c|c|c|c|c|c|}
\hline
Dataset & Method & Feature & Model & Accuracy & AUC & Recall & Prec. & F1 \\ \hline
\multirow{8}{*}{\begin{tabular}[c]{@{}c@{}}Liver\\ Tumours\end{tabular}}
& \begin{tabular}[c]{@{}c@{}}Cubical Complex \\ \& Barcod Agg.\end{tabular} & Pers. Statistics & LDA & 0.8250 & 0.8875 & 0.8250 & 0.8258 & 0.8249 \\ \cline{2-9} 
& \begin{tabular}[c]{@{}c@{}}Cubical Complex \\ \& Feature Concat.\end{tabular} & Betti Curve & LightGBM & \textbf{0.9250} & \textbf{0.9375} & \textbf{0.9250} & \textbf{0.9261} & \textbf{0.9250} \\ \cline{2-9}
& \begin{tabular}[c]{@{}c@{}}$VR$ Complex \\ \& Barcod Agg.\end{tabular} & Pers. Statistics & CatBoost & 0.7500 & \textbf{0.8200} & 0.7500 & 0.7525 & 0.7494 \\ \cline{2-9} 
& \begin{tabular}[c]{@{}c@{}}$VR$ Complex \\ \& Feature Concat.\end{tabular} & \begin{tabular}[c]{@{}c@{}}Entropy \\Summary\end{tabular} & RF & \textbf{0.8000} & 0.7975 & \textbf{0.8000} & \textbf{0.8000} & \textbf{0.8000} \\
\hline

\multirow{8}{*}{\begin{tabular}[c]{@{}c@{}}KiTS19\end{tabular}}
& \begin{tabular}[c]{@{}c@{}}Cubical Complex \\ \& Barcod Agg.\end{tabular} & \begin{tabular}[c]{@{}c@{}}Entropy \\Summary\end{tabular} & LR & 0.7750 & 0.7198 & 0.6429 & \textbf{0.6923} & 0.6667 \\ \cline{2-9} 
& \begin{tabular}[c]{@{}c@{}}Cubical Complex \\ \& Feature Concat.\end{tabular} & Betti Curve & ET & \textbf{0.8000} & \textbf{0.8104} & \textbf{0.7857} & 0.6875 & \textbf{0.7333} \\ \cline{2-9}
& \begin{tabular}[c]{@{}c@{}}$VR$ Complex \\ \& Barcod Agg.\end{tabular} & \begin{tabular}[c]{@{}c@{}}Entropy \\Summary\end{tabular} & Ridge & \textbf{0.8250} & \textbf{0.8159} & \textbf{0.7857} & 0.7333 & \textbf{0.7586} \\ \cline{2-9} 
& \begin{tabular}[c]{@{}c@{}}$VR$ Complex \\ \& Feature Concat.\end{tabular} & P.T.C. & KNN & \textbf{0.8250} & 0.8132 & 0.7143 & \textbf{0.7692} & 0.7407 \\
\hline
\end{tabular}
\end{center}
\end{table}

In the second half of table \ref{tab:test_matrics_3D_datasets}, the results of the experiments on the KiTS19 datasets are reported for both cubical complex and landmark-based $VR$ complex filtration methods. As demonstrated in the table, feature concatenation outperforms barcode aggregation in terms of all the metrics when cubical complex filtration is used; however, both approaches yield analogous results when barcodes are constructed via landmark-based $VR$ complex filtration. Additional experimental results on the KiTS19 dataset, including the results of the best performing models with all five vectorization methods, can be found in tables \ref{tab:test_matrics_KiTS19_dataset_Cubical_Complex} and \ref{tab:test_matrics_KiTS19_dataset_ULBP}.

The experimental results on the 2D datasets, BUSI and DDSM, are depicted in table \ref{tab:test_matrics_2D_datasets} where we only used $VR$ with ULBP G4 and G5 including all of their 8 rotations. P.T.C. features trained with the CatBoost classifier obtained the best performance with an F1-score of 0.8974 and AUC of 0.9282 based on the feature concatenation approach. Results of the experiment on the DDSM dataset demonstrate similar trends where feature concatenation is the dominant method in terms of all the evaluation metrics. Detailed experimental results of the 2D datasets with other vectorization methods can be found in tables \ref{tab:test_matrics_BUSI_dataset_ULBP} and \ref{tab:test_matrics_DDSM_dataset_ULBP} for BUSI and DDSM datasets, respectively.

Across all the experimental results of the 2D and 3D datasets with various filtration methods, feature concatenation outperformed the barcode aggregation approach in terms of almost all performance metrics except in one instance, see table \ref{tab:test_matrics_3D_datasets} where similar results were drawn from both methods when $VR$ deployed and metrics show no statistically significant differences between barcode aggregation with vectorized barcode concatenation. Albeit the variation in precision and recall among the two methods, one can see the F1 score as a balanced measure of both metrics especially in unbalanced scenarios as in KiTS19 dataset. These findings suggest that the former approach should be prioritized when experimenting on datasets of similar modalities having more than a barcode constructed from a single image.

\begin{table}[!h]
\caption{Comparison of Barcode Aggregation and Feature Concatenation Approach Based on Their Best Performing Classifier and Feature Type on the Testing Data of 2D Datasets Using Landmark-Based $VR$ Complex Filtration.}
\label{tab:test_matrics_2D_datasets}
\begin{center}
\begin{tabular}{|c|c|c|c|c|c|c|c|c|}
\hline
Dataset & Method & Feature & Model & Accuracy & AUC & Recall & Prec. & F1 \\ \hline
\multirow{4}{*}{\begin{tabular}[c]{@{}c@{}}BUSI\end{tabular}}
& \begin{tabular}[c]{@{}c@{}}$VR$ Complex \\ \& Barcod Agg.\end{tabular} & Betti Curve & CatBoost & 0.8333 & 0.8650 & \textbf{0.9722} & 0.7609 & 0.8537 \\ \cline{2-9} 
& \begin{tabular}[c]{@{}c@{}}$VR$ Complex \\ \& Feature Concat.\end{tabular} & P.T.C. & CatBoost & \textbf{0.8889} & \textbf{0.9282} & \textbf{0.9722} & \textbf{0.8333} & \textbf{0.8974} \\
\hline

\multirow{4}{*}{\begin{tabular}[c]{@{}c@{}}DDSM\end{tabular}}
& \begin{tabular}[c]{@{}c@{}}$VR$ Complex \\ \& Barcod Agg.\end{tabular} & Pers. Statistics & LR & 0.8039 & 0.8900 & 0.7843 & 0.8163 & 0.8000 \\ \cline{2-9} 
& \begin{tabular}[c]{@{}c@{}}$VR$ Complex \\ \& Feature Concat.\end{tabular} & Pers. Landscape  & Ensemble & \textbf{0.8529} & \textbf{0.9212} & \textbf{0.8627} & \textbf{0.8462} & \textbf{0.8544} \\
\hline
\end{tabular}
\end{center}
\end{table}

\section{Conclusion}
In this study, we compared two methods of combining multiple barcodes: aggregation and concatenation. The results of this study inform the usage of topological features within ML pipelines from a single 2D image or multiple slices of 3D CT scan. Specifically, the effects of persistence barcode aggregation were compared to vectorized barcode feature concatenation as two different approaches to summarize topological and geometrical features from multiple barcodes constructed from a single 2D or 3D medical image. Experiments were conducted on four medical imaging datasets: two datasets of CT scans, one ultrasound and one mammography dataset. This is together with two filtration methods to generate barcodes: cubical complex and landmark-based $VR$ complex filtration. A comprehensive classification pipeline was used with five different barcode vectorization approaches to benchmark and draw a fair comparison of the two approaches. Our findings suggest that, overall, fusing feature vectors computed from multiple PBs results in better performing models than a feature vector built from the aggregation of multiple PBs. These findings can serve as a useful guideline for TDA practitioners across domains conducting similar analysis and for researchers in other fields looking to apply TDA in their research. One limitation of this work is the lack of inclusion of more medical imaging datasets and other medical imaging modalities such as magnetic resonance imaging (MRI) or histopathological data.
Future research directions include exploring other approaches to building PH from medical images and incorporating other clinical variations, such as noise index or slice thickness in CT images, for instance, into the experiments conducted in this study.

\section*{Acknowledgments}       
This work was funded by National Institutes of Health and National Cancer Institute grants R01CA233888 and U01CA238444.

\clearpage
\appendix
\section{Supplementary Tables}

\begin{table}[!h]
\caption{Comparison of Barcode Aggregation with Feature concatenation Approach Based on Their Best Performing Classifier for Each Feature Type on the Testing Data of Liver Tumours Dataset Using Cubical Complex Filtration.}
\label{tab:test_matrics_HCC_ICC_dataset_cubical_conplex}
\begin{center}
\begin{tabular}{|c|c|c|c|c|c|c|c|}
\hline
Method                                                                              & Feature          & Model               & Accuracy        & AUC             & Recall          & Prec.           & F1              \\ \hline
\multirow{5}{*}{\begin{tabular}[c]{@{}c@{}}Barcod\\ Agg.\end{tabular}}              & Betti Curve      & RF       & 0.7000          & 0.7450          & 0.7000          & 0.7020          & 0.6992          \\ \cline{2-8} 
& Entropy Summary  & XGBoost      & 0.6000          & 0.6825          & 0.6000          & 0.6042          & 0.5960          \\ \cline{2-8} 
& Pers. Statistics & LDA                 & 0.8250          & 0.8875          & 0.8250          & 0.8258          & 0.8249          \\ \cline{2-8} 
& Pers. Landscape  & LightGBM                & 0.7000          & 0.7525          & 0.7000          & 0.7083          & 0.6970          \\ \cline{2-8} 
& P.T.C.       & ET         & 0.6000          & 0.6575          & 0.6000          & 0.6010          & 0.5990          \\ \hline
\multirow{5}{*}{\textbf{\begin{tabular}[c]{@{}c@{}}Feature\\ Concat.\end{tabular}}} & Betti Curve      & LightGBM                & \textbf{0.9250} & \textbf{0.9375} & \textbf{0.9250} & \textbf{0.9261} & \textbf{0.9250} \\ \cline{2-8} 
& Entropy Summary  & RF       & 0.8500          & 0.8800          & 0.8500          & 0.8535          & 0.8496          \\ \cline{2-8} 
& Pers. Statistics & LR & 0.7250          & 0.8375          & 0.7250          & 0.7256          & 0.7248          \\ \cline{2-8} 
& Pers. Landscape  & CatBoost           & 0.7250          & 0.8275          & 0.7250          & 0.7302          & 0.7234          \\ \cline{2-8} 
& P.T.C.       & CatBoost           & 0.7750          & 0.8775          & 0.7750          & 0.7757          & 0.7749          \\ \hline
\end{tabular}
\end{center}
\end{table}

\begin{table}[!h]
\caption{Comparison of Barcode Aggregation with Feature Concatenation Approach Based on Their Best Performing Classifier for Each Feature Type on the Testing Data of the Liver Tumours Dataset Using Landmark-Based $VR$ Complex Filtration.}
\label{tab:test_matrics_HCC_ICC_dataset_ULBP}
\begin{center}
\begin{tabular}{|c|c|c|c|c|c|c|c|}
\hline
Method                                                                              & Feature          & Model             & Accuracy        & AUC             & Recall          & Prec.           & F1              \\ \hline
\multirow{5}{*}{\begin{tabular}[c]{@{}c@{}}Barcod\\ Agg.\end{tabular}}              & Betti Curve      & RF     & 0.7250          & \textbf{0.8438} & 0.7250          & 0.7564          & 0.7163          \\ \cline{2-8} 
& Entropy Summary  & ET       & 0.7000          & 0.7200          & 0.7000          & 0.7083          & 0.6970          \\ \cline{2-8} 
& Pers. Statistics & CatBoost         & 0.7500          & 0.8200          & 0.7500          & 0.7525          & 0.7494          \\ \cline{2-8} 
& Pers. Landscape  & LDA               & 0.5750          & 0.6225          & 0.5750          & 0.5800          & 0.5683          \\ \cline{2-8} 
& P.T.C.       & RF     & 0.6000          & 0.5075          & 0.6000          & 0.6099          & 0.5908          \\ \hline
\multirow{5}{*}{\textbf{\begin{tabular}[c]{@{}c@{}}Feature\\ Concat.\end{tabular}}} & Betti Curve      & LR                & 0.7250          & 0.7250          & 0.7250          & 0.7302          & 0.7234          \\ \cline{2-8} 
& Entropy Summary  & RF     & \textbf{0.8000} & 0.7975          & \textbf{0.8000} & \textbf{0.8000} & \textbf{0.8000} \\ \cline{2-8} 
& Pers. Statistics & XGBoost               & 0.6750          & 0.6950          & 0.6750          & 0.6754          & 0.6748          \\ \cline{2-8} 
& Pers. Landscape  & GBC & 0.7500          & 0.8175          & 0.7500          & 0.7525          & 0.7494          \\ \cline{2-8} 
& P.T.C.       & CatBoost         & 0.6250          & 0.6850          & 0.6250          & 0.6279          & 0.6229          \\ \hline
\end{tabular}
\end{center}
\end{table}

\begin{table}[!h]
\caption{Comparison of Barcode Aggregation with Feature Concatenation Approach Based on Their Best Performing Classifier for Each Feature Type on the Testing Data of the KiTS19 Dataset Using Cubical Complex Filtration.}
\label{tab:test_matrics_KiTS19_dataset_Cubical_Complex}
\begin{center}
\begin{tabular}{|c|c|c|c|c|c|c|c|}
\hline
Method                                                                              & Feature          & Model         & Accuracy        & AUC             & Recall          & Prec.           & F1              \\ \hline
\multirow{5}{*}{\begin{tabular}[c]{@{}c@{}}Barcod\\ Agg.\end{tabular}}              & Betti Curve      & RF & 0.7000          & 0.7418          & 0.5714          & 0.5714          & 0.5714          \\ \cline{2-8} 
& Entropy Summary  & LR            & 0.7750          & 0.7198          & 0.6429          & 0.6923          & 0.6667          \\ \cline{2-8} 
& Pers. Statistics & Ridge         & 0.7500          & 0.7253          & 0.6429          & 0.6429          & 0.6429          \\ \cline{2-8} 
& Pers. Landscape  & QDA           & 0.7500          & 0.6841          & 0.6429          & 0.6429          & 0.6429          \\ \cline{2-8} 
& P.T.C.       & RF & 0.7250          & 0.7692          & 0.7143          & 0.5882          & 0.6452          \\ \hline
\multirow{5}{*}{\textbf{\begin{tabular}[c]{@{}c@{}}Feature\\ Concat.\end{tabular}}} & Betti Curve      & ET   & 0.8000          & 0.8104          & \textbf{0.7857} & 0.6875          & \textbf{0.7333} \\ \cline{2-8} 
& Entropy Summary  & CatBoost     & 0.8000          & 0.8049          & 0.6429          & 0.7500          & 0.6923          \\ \cline{2-8} 
& Pers. Statistics & Ensemble      & \textbf{0.8250} & \textbf{0.8159} & 0.6429          & \textbf{0.8182} & 0.7200          \\ \cline{2-8} 
& Pers. Landscape  & LR            & 0.8000          & 0.7940          & 0.7143          & 0.7143          & 0.7143          \\ \cline{2-8} 
& P.T.C.       & LR            & 0.7750          & 0.7500          & 0.7143          & 0.6667          & 0.6897          \\ \hline
\end{tabular}
\end{center}
\end{table}

\begin{table}[!h]
\caption{Comparison of Barcode Aggregation with Feature Concatenation Approach Based on Their Best Performing Classifier for Each Feature Type on the Testing Data of the KiTS19 Dataset Using Landmark-Based $VR$ Complex Filtration.}
\label{tab:test_matrics_KiTS19_dataset_ULBP}
\begin{center}
\begin{tabular}{|c|c|c|c|c|c|c|c|}
\hline
Method                                                                              & Feature          & Model         & Accuracy        & AUC             & Recall          & Prec.           & F1              \\ \hline
\multirow{5}{*}{\textbf{\begin{tabular}[c]{@{}c@{}}Barcod\\ Agg.\end{tabular}}}     & Betti Curve      & ET   & 0.7750          & 0.7637          & 0.7143          & 0.6667          & 0.6897          \\ \cline{2-8} 
& Entropy Summary  & Ridge         & \textbf{0.8250} & 0.8159          & \textbf{0.7857} & 0.7333          & \textbf{0.7586} \\ \cline{2-8} 
& Pers. Statistics & Ensemble      & \textbf{0.8250} & 0.8077          & \textbf{0.7857} & 0.7333          & \textbf{0.7586} \\ \cline{2-8} 
& Pers. Landscape  & LightGBM          & 0.8000          & 0.8255          & 0.6429          & 0.7500          & 0.6923          \\ \cline{2-8} 
& P.T.C.       & SVM           & 0.7250          & 0.7060          & 0.6429          & 0.6000          & 0.6207          \\ \hline
\multirow{5}{*}{\textbf{\begin{tabular}[c]{@{}c@{}}Feature\\ Concat.\end{tabular}}} & Betti Curve      & CatBoost     & 0.7500          & 0.7610          & 0.5714          & 0.6667          & 0.6154          \\ \cline{2-8} 
& Entropy Summary  & LR            & 0.7500          & 0.7720          & 0.6429          & 0.6429          & 0.6429          \\ \cline{2-8} 
& Pers. Statistics & ADA     & 0.8000          & \textbf{0.8379} & \textbf{0.7857} & 0.6875          & 0.7333          \\ \cline{2-8} 
& Pers. Landscape  & RF & 0.7250          & 0.8077          & 0.6429          & 0.6000          & 0.6207          \\ \cline{2-8} 
& P.T.C.       & KNN   & \textbf{0.8250} & 0.8132          & 0.7143          & \textbf{0.7692} & 0.7407          \\ \hline
\end{tabular}
\end{center}
\end{table}

\begin{table}[!h]
\caption{Comparison of Barcode Aggregation with Feature Concatenation Approach Based on Their Best Performing Classifier for Each Feature Type on the Testing Data of the BUSI Dataset Using Landmark-Based $VR$ Complex Filtration.}
\label{tab:test_matrics_BUSI_dataset_ULBP}
\begin{center}
\begin{tabular}{|c|c|c|c|c|c|c|c|}
\hline
Method                                                                              & Feature             & Model              & Accuracy        & AUC             & Recall          & Prec.           & F1              \\ \hline
\multirow{5}{*}{\begin{tabular}[c]{@{}c@{}}Barcod\\ Agg.\end{tabular}}              & Betti Curve         & CatBoost          & 0.8333          & 0.8650          & \textbf{0.9722} & 0.7609          & 0.8537          \\ \cline{2-8} 
& Entropy Summary     & LR                 & 0.8056          & 0.8850          & 0.9167          & 0.7500          & 0.8250          \\ \cline{2-8} 
& Pers. Statistics    & CatBoost          & 0.8333          & 0.9051          & 0.9167          & 0.7857          & 0.8462          \\ \cline{2-8} 
& Pers. Landscape     & Ensemble           & 0.7361          & 0.8071          & 0.8333          & 0.6977          & 0.7595          \\ \cline{2-8} 
& P.T.C.          & CatBoost          & 0.7500          & 0.8079          & 0.8611          & 0.7045          & 0.7750          \\ \hline
\multirow{5}{*}{\textbf{\begin{tabular}[c]{@{}c@{}}Feature\\ Concat.\end{tabular}}} & Betti Curve         & RF      & 0.8750          & 0.8758          & 0.9444          & 0.8293          & 0.8831          \\ \cline{2-8} 
& Entropy Summary     & GBC  & 0.8472          & 0.9136          & 0.9167          & 0.8049          & 0.8571          \\ \cline{2-8} 
& Pers. Statistics    & CatBoost          & 0.8194          & 0.9128          & 0.8889          & 0.7805          & 0.8312          \\ \cline{2-8} 
& Pers. Landscape     & ET        & 0.8056          & 0.8939          & 0.9444          & 0.7391          & 0.8293          \\ \cline{2-8} 
& P.T.C.          & CatBoost          & \textbf{0.8889} & \textbf{0.9282} & \textbf{0.9722} & \textbf{0.8333} & \textbf{0.8974} \\ \hline
\end{tabular}
\end{center}
\end{table}

\begin{table}[!h]
\caption{Comparison of Barcode Aggregation with Feature Concatenation Approach Based on Their Best Performing Classifier for Each Feature Type on the Testing Data of the DDSM Dataset Using Landmark-Based $VR$ Complex Filtration.}
\label{tab:test_matrics_DDSM_dataset_ULBP}
\begin{center}
\begin{tabular}{|c|c|c|c|c|c|c|c|}
\hline
Method                                                                              & Feature          & Model             & Accuracy        & AUC             & Recall          & Prec.           & F1              \\ \hline
\multirow{5}{*}{\begin{tabular}[c]{@{}c@{}}Barcod\\ Agg.\end{tabular}}              & Betti Curve      & GBC & 0.7941          & 0.8216          & 0.6863          & 0.8750          & 0.7692          \\ \cline{2-8} 
& Entropy Summary  & ET       & 0.7451          & 0.8541          & 0.6863          & 0.7778          & 0.7292          \\ \cline{2-8} 
& Pers. Statistics & LR                & 0.8039          & 0.8900          & 0.7843          & 0.8163          & 0.8000          \\ \cline{2-8} 
& Pers. Landscape  & Ensemble          & 0.7745          & 0.8460          & 0.7843          & 0.7692          & 0.7767          \\ \cline{2-8} 
& P.T.C.       & LR                & 0.8039          & 0.8939          & 0.7647          & 0.8298          & 0.7959          \\ \hline
\multirow{5}{*}{\textbf{\begin{tabular}[c]{@{}c@{}}Feature\\ Concat.\end{tabular}}} & Betti Curve      & Ensemble          & 0.8431          & 0.8964          & 0.8431          & 0.8431          & 0.8431          \\ \cline{2-8} 
& Entropy Summary  & Ensemble          & 0.8137          & 0.8927          & 0.7647          & 0.8478          & 0.8041          \\ \cline{2-8} 
& Pers. Statistics & CatBoost         & 0.8529          & \textbf{0.9319} & 0.8039          & 0.8913          & 0.8454          \\ \cline{2-8} 
& Pers. Landscape  & Ensemble          & 0.8529          & 0.9212          & \textbf{0.8627} & 0.8462          & \textbf{0.8544} \\ \cline{2-8} 
& P.T.C.       & RF     & \textbf{0.8627} & 0.8997          & 0.7843          & \textbf{0.9302} & 0.8511          \\ \hline
\end{tabular}
\end{center}
\end{table}

\bibliographystyle{unsrt}  
\bibliography{references}

\end{document}